\documentclass[sigconf]{aamas} 
\usepackage{multirow}
\usepackage{balance} % for balancing columns on the final page

\setcopyright{ifaamas}
\acmConference[AAMAS '24]{Proc.\@ of the 23rd International Conference
on Autonomous Agents and Multiagent Systems (AAMAS 2024)}{May 6 -- 10, 2024}
{Auckland, New Zealand}{N.~Alechina, V.~Dignum, M.~Dastani, J.S.~Sichman (eds.)}
\copyrightyear{2024}
\acmYear{2024}
\acmDOI{}
\acmPrice{}
\acmISBN{}

\acmSubmissionID{<<EasyChair submission id>>}

\title[PDiT: Interleaving Perception and Decision Transformer for DRL]{PDiT: Interleaving Perception and Decision-making Transformers for Deep Reinforcement Learning}

\author{Hangyu Mao}
\affiliation{
  \institution{SenseTime Research}
  \city{Beijing}
  \country{China}}
\email{maohangyu@sensetime.com}

\author{Rui Zhao}
\affiliation{
  \institution{SenseTime Research}
  \city{Shenzhen}
  \country{China}}
\email{zhaorui@sensetime.com}

\author{Ziyue Li}
\affiliation{
  \institution{University of Cologne}
  \city{Cologne}
  \country{Germany}}
\email{zlibn@wiso.uni-koeln.de}

\author{Zhiwei Xu}
\affiliation{
  \institution{University of Chinese Academy of Sciences}
  \city{Beijing}
  \country{China}}
\email{xuzhiwei2019@ia.ac.cn}

\author{Hao Chen}
\affiliation{
  \institution{University of Chinese Academy of Sciences}
  \city{Beijing}
  \country{China}}
\email{chenhao915@mails.ucas.ac.cn}

\author{Yiqun Chen}
\affiliation{
  \institution{Gaoling School of AI, Renmin University of China}
  \city{Beijing}
  \country{China}}
\email{chenyiqun2020@ia.ac.cn}

\author{Bin Zhang}
\affiliation{
  \institution{University of Chinese Academy of Sciences}
  \city{Beijing}
  \country{China}}
\email{zhangbin2020@ia.ac.cn}

\author{Zhen Xiao}
\affiliation{
  \institution{Peking University}
  \city{Beijing}
  \country{China}}
\email{xiaozhen@pku.edu.cn}

\author{Junge Zhang}
\affiliation{
  \institution{Institute of Automation, Chinese Academy of Sciences}
  \city{Beijing}
  \country{China}}
\email{jgzhang@nlpr.ia.ac.cn}

\author{Jiangjin Yin}
\affiliation{
  \institution{Huazhong Agricultural University}
  \city{Wuhan}
  \country{China}}
\email{jiangjinyin@mail.hzau.edu.cn}

\begin{abstract}
Designing better deep networks and better reinforcement learning (RL) algorithms are both important for deep RL. This work studies the former. Specifically, the Perception and Decision-making Interleaving Transformer (PDiT) network is proposed, which cascades two Transformers in a very natural way: the perceiving one focuses on \emph{the environmental perception} by processing the observation at the patch level, whereas the deciding one pays attention to \emph{the decision-making} by conditioning on the history of the desired returns, the perceiver's outputs, and the actions. Such a network design is generally applicable to a lot of deep RL settings, e.g., both the online and offline RL algorithms under environments with either image observations, proprioception observations, or hybrid image-language observations. Extensive experiments show that PDiT can not only achieve superior performance than strong baselines in different settings but also extract explainable feature representations. Our code is available at \url{https://github.com/maohangyu/PDiT}.
\end{abstract}


\newcommand{\BibTeX}{\rm B\kern-.05em{\sc i\kern-.025em b}\kern-.08em\TeX}

\begin{document}

%%% The following commands remove the headers in your paper. For final 
%%% papers, these will be inserted during the pagination process.

\pagestyle{fancy}
\fancyhead{}

%%% The next command prints the information defined in the preamble.

\maketitle 

%%%%%%%%%%%%%%%%%%%%%%%%%%%%%%%%%%%%%%%%%%%%%%%%%%%%%%%%%%%%%%%%%%%%%%%%

\section{Introduction}
Deep reinforcement learning (RL) has made great progress in automating decisions in complex environments and tasks, from virtual environments such as mastering video games \cite{mnih2015human} to real-world applications such as object grasp, autonomous driving, and 6-DoF manipulation \cite{evans2016deepmind}. These environments and tasks usually involve multi-modal data. 

\begin{figure}[t]
\centering
\includegraphics[width=0.96\columnwidth]{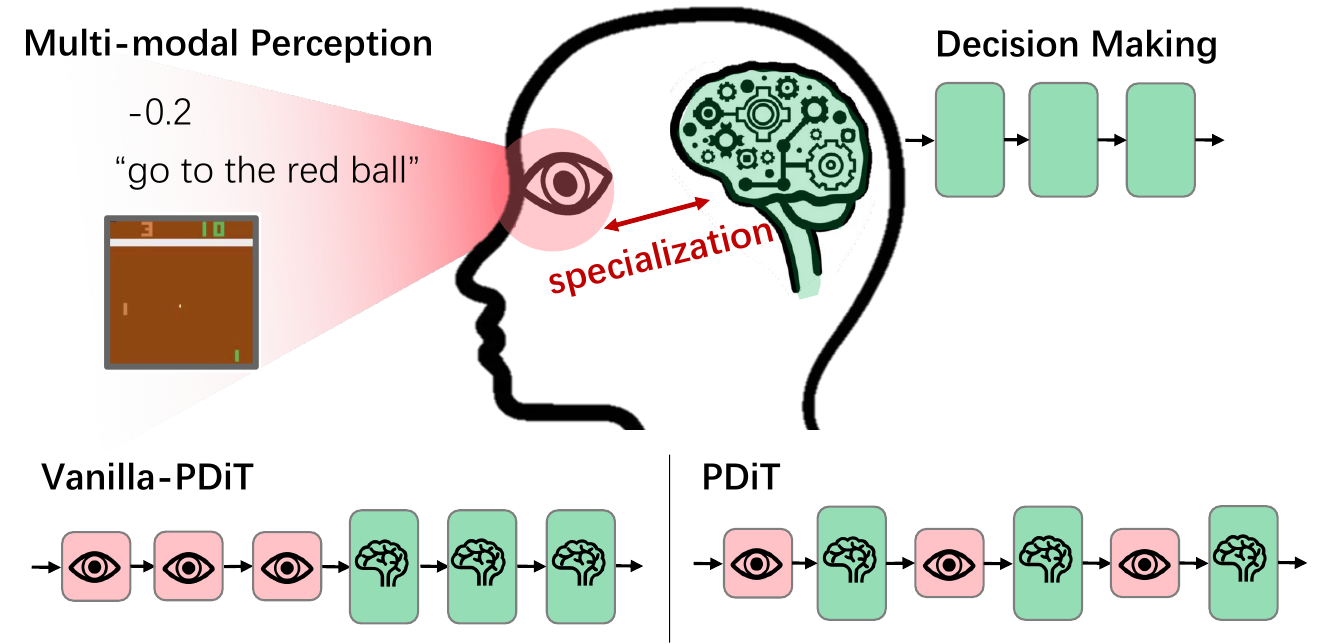}
\caption{Perception and Decision Making.}
\label{fig:pdit}
\end{figure}

Generally, an RL agent receives the environmental observations (\textbf{perception}) and takes actions accordingly (\textbf{decision-making}) to maximize the accumulated rewards. Dealing with environments with multi-modal data can be rather challenging, and various deep learning modules have been employed in perception and decision-making to enhance performance. 

For most of the deep RL models to perceive inputs in a specific modality from the environment, the perception module needs to be chosen accordingly. Commonly, Multi-Layer Perceptron (MLP), Convolutional Neural Network (CNN), and Recurrent Neural Network (RNN) with its variances have been adopted to encode low-dimensional or proprioception observation \cite{hafner2023mastering}, vision \cite{mnih2015human, cai2023open}, and language \cite{hausknecht2015deep}, respectively. It will be more challenging when the environment contains more than one modality. The intuitive solution was combining different perception modules. For instance, DreamerV3 \cite{hafner2023mastering} uses a combination of MLPs and CNN for a mixture of low-dimensional and image inputs. However, such a combination inevitably complicates the models.

From the perspective of decision-making, deep modules such as MLP are commonly used to parameterize the policy, e.g., in classic RL models such as DQN \cite{mnih2015human} and Actor-Critic \cite{John2017PPO}. Given the nature of RL as a sequential decision process, Transformer \cite{vaswani2017attention} has also recently been adopted, proving that it is a powerful decision-maker. 

However, when considering perception and decision-making together in multi-modal environments, challenges emerge when designing the model: an intuitive solution could still be stacking various modules together. We denote Perception and Decision-making as P and D for short: For example, CoBERL \cite{banino2021coberl} use residual network (P) and Transformer (D); Catformer \cite{davis2021catformer} uses CNN (P) and Transformer (D); when dealing with vision-language tasks, RT-1 \cite{brohan2022rt} uses residual network for vision (P), Universal-Sentence-Encoder for language (D), and Transformer for action (D). The models tend to be gradually over-designed and complicated.

In parallel, owing to the Transformer's growing ability in vision-language domains, such as Vision Transformer \cite{Arnab2021ViViT}, Uni-Perceiver-v2 \cite{li2023uni}, PaLM-E \cite{driess2023palm}, it has been proved that Transformer has powerful vision-language multi-modal perception. This development largely encourages researchers to simplify the model design for a multi-modal generalist agent. Ground-breaking offline RL works such as Decision Transformer (DT) \cite{chen2021decision}, Trajectory Transformer (TT) \cite{janner2021reinforcement}, Gato \cite{reed2022generalist}, and MGDT \cite{lee2022multi} only use a single Transformer sequence model: it encodes inputs from various tasks in different modalities with a causal Transformer, and predict the next step's action with the same Transformer in an autoregressive manner. Undoubtedly, this simple design eases the model complexity; however, \textbf{fusing perception and decision-making in the same Transformer model could only be sub-optimal}. Instead, delegating perception and decision to two separate modules is intuitively more efficient, and \citet{stooke2021decoupling} also proves that a specialized perceiver and a specialized decision-maker could largely accelerate learning. At the same time, a compact model design is still preferred over the module-stacking models. 

To this end, we propose the Perception and Decision-making Interleaving Transformer (PDiT) network, which specializes in perception and decision with two respective Transformers as shown in Figure \ref{fig:pdit}. Specifically, the PDiT network is made up of a Perceiving Transformer (perceiver) and a Deciding Transformer (decision-maker): the perceiver processes the observation at the patch level to learn a \emph{good environmental understanding}, while the decision-maker focuses on the \emph{good policy learning} by conditioning on the history of the desired returns, the perceiver's outputs and the actions. We first propose a \textit{Vanilla-PDiT}, which naively stacks the Deciding Transformer on top of the Perceiving Transformer. 

However, our experiments find that Vanilla-PDiT cannot always achieve the expected performance. Possible reasons are that the vanilla version has no information exchange between the perception and decision; besides, stacking several Transformer blocks may cause over-global attention and loss of focus \cite{dosovitskiy2020image, ViTRepresentation}. Since the deciding layer is to make decisions, once the hidden deciding layer receives the signal from the perceiving layer, it should directly take actions instead of further embedding the input’s representations abstractly by stacking multiple deciding layers.

Thus, we propose the final model, which builds a PDiT block by interleaving the perceiving and deciding Transformer blocks and then stacks $L$ PDiT blocks to form the network. In this way, both perception and decision are considered, and they are still specialized from the perspective of each PDiT block, and the performance is better in practice. Our contributions are summarized:
\begin{itemize}
    \item The core contribution is the proposed PDiT network (Vanilla-PDiT and full PDiT), which is generally applicable to various deep RL settings: (1) \textbf{online and offline}: the online RL like PPO \cite{John2017PPO}, the offline RL like CQL \cite{kumar2020conservative}, and the offline supervised learning like DT; (2) \textbf{multi-modal inputs}: the environments with \textit{image} observations like Atari, the environments with \textit{proprioception} observations like MuJoCo, and the environments with \textit{hybrid image-language} observations like BabyAI; (3) \textbf{multi-task reinforcement learning} without changing the network architecture.
    \item Furthermore, extensive experiments show that the full PDiT can achieve superior performance than strong baselines in different settings. Ablation study and feature visualization give a better understanding of our methods.
\end{itemize}

\section{Related Work}
\begin{figure}[t]
\centering
\includegraphics[width=1.0\columnwidth]{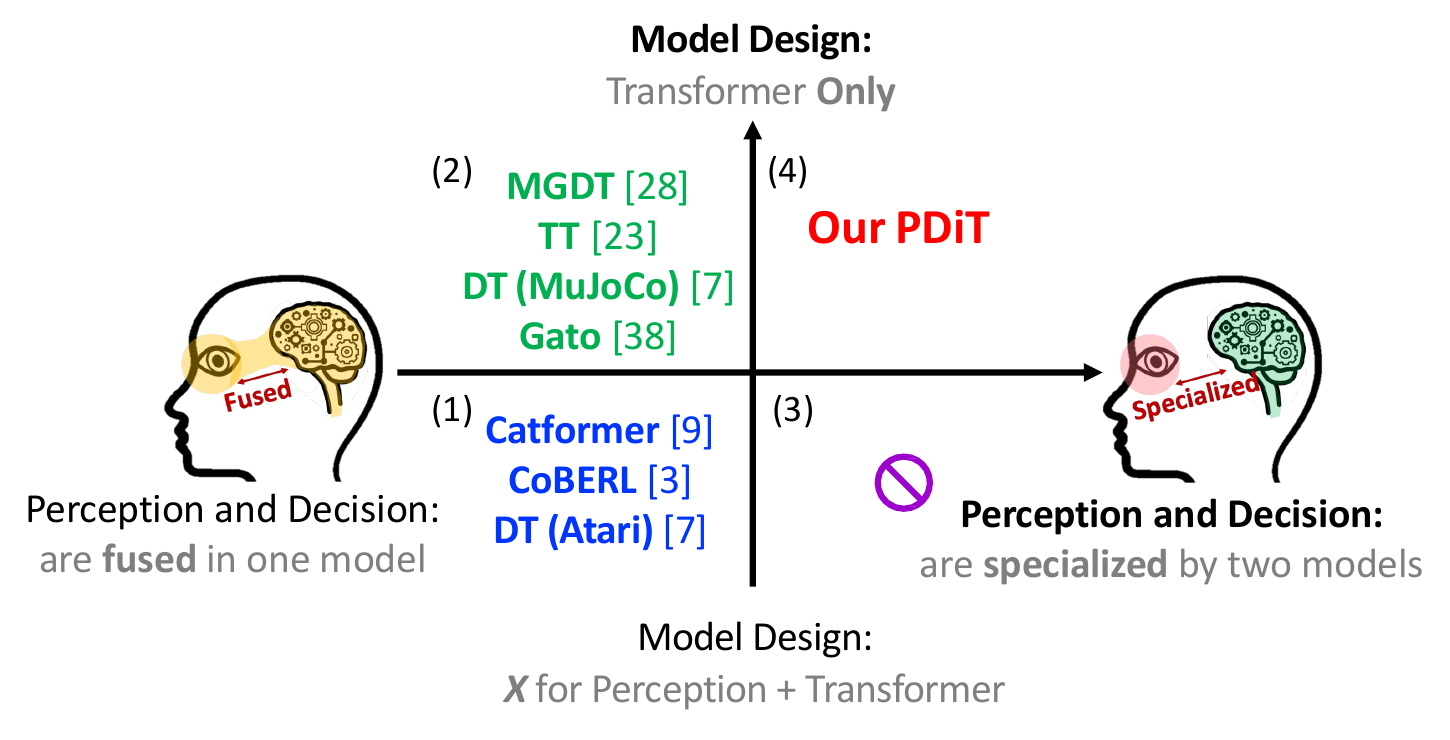}
\caption{The categories of Transformer-based RL methods, in terms of function specialization (i.e., $x$-axis) and model design (i.e., $y$-axis). Our PDiT combines the advantages of recent methods. This illustration does not include all related works for simplicity.}
\label{fig:CurrentTRL}
\end{figure}

\textbf{A Landscape.} Recently, there is a trend of applying Transformers to deep RL \cite{hu2022TRLsurvey,li2023TRLsurvey,yuan2023transformer}. As shown in Figure \ref{fig:CurrentTRL}, we categorize the Transformer-based methods in two dimensions: a) Function specialization: whether the two essential functions of environmental perception and decision-making are fused in one model or specialized with two models; b) Model architecture: whether it is \textit{Transformer-only} or \textit{X+Transformer}, which combines a Transformer with other deep $X$ modules such as CNN to perceive the environment. Specifically, we have the following four combinations: (1) \textcolor{blue}{\textit{Specialized: X+Transformer}} methods such as GTrXL \cite{parisotto2020stabilizing}, Catformer \cite{davis2021catformer}, CoBERL \cite{banino2021coberl}, DT  \cite{chen2021decision} in Atari environment. For instance, Catformer \cite{davis2021catformer} combines CNN with Transformer, where the CNN is specialized for perception, and the Transformer is specialized for decision.  In practice, they need to change their perception module when handling different observation types, which may hinder their scalability in complex environments. (2) In contrast, the \textcolor{Green}{\textit{Fused: Transformer-only}} methods like MGDT \cite{lee2022multi}, TT \cite{janner2021reinforcement}, DT in MuJuCo environment, Gato \cite{reed2022generalist}, are purely based on the Transformer, which fuses the role of environmental perception and decision-making simultaneously in one Transformer. These models significantly simplified the model structure and improved the scalability, but such a perception-decision fusion in the same model could be sub-optimal and slow in training \cite{stooke2021decoupling}. For example, DT \cite{chen2021decision} mixes environmental observations with returns and actions in one Transformer when testing on the MuJoCo tasks, but the performance can be further improved by processing observations separately with another Transformer, as shown by our experiments. (3) There are no \textcolor{Plum}{\textit{Fused: Transformer+X}}, which could be reasonable due to them being potentially over-complicated. (4) The proposed PDiT is, to the best of our knowledge, \textbf{the first model that delegates perception and decision-making into two Transformers}, leveraging Transformer's power of dealing with multi-modal data and making sequential decisions.

\textbf{Concurrent Work.} StARformer \cite{shang2022starformer}, HDT \cite{correia2022hierarchical}, GDT \cite{hu2023graph} and our preliminary work TIT \cite{mao2022transformer} also apply two Transformers for better scalability and performance. But we have fundamentally different motivations and implementations: StARformer focuses on balancing short-term and long-term dependencies with mixed Transformer-CNN; HDT aims at hierarchical control with subgoal prediction; GDT applies a causal graph to capture potential dependencies between different concepts and facilitate temporal and causal relationship learning; in contrast, our PDiT combines the advantages of the existing methods by dividing environmental perception and decision-making into two Transformers, and it is applicable for many deep RL settings.

\textbf{Applying Two Transformers in Other Fields.} In multi-agent reinforcement learning, two Transformers are used for asynchronous action coordination \cite{zhang2023stackelberg}. 
In computer vision, some methods apply two Transformers to handle tasks like image classification and person search, e.g., TNT \cite{han2021transformer}., ViViT \cite{Arnab2021ViViT}, DualFormer \cite{Liang2022DualFormer} and COAT \cite{Yu2022CascadeTransformers}. However, our PDiT and these works are totally different. Although both divide the image input into patches, it is worth noting that this is a standard process for all Vision Transformers. Besides, both have two Transformers but with different designs. For example, TNT’s outer Transformer takes the coarse-grain patches, and the perceiving one further takes the finer patches from the coarse patch. For our PDiT, the perceiver’s input is the multi-modal observation (image, proprioception, image-language), and the decision-maker’s input is return-observation-action embedding. The perceiving Transformer perceives the environment, and the deciding one takes action. As a result, TNT can only be used for vision tasks, but our PDiT is designed for RL tasks.

\section{Approach}
We consider problems that can be formulated as a Markov Decision Process (MDP), which is formally defined by a tuple $\langle S, A, T, R, \gamma \rangle$, where $S$ is the set of possible states $s \in S$; $A$ represents the set of possible actions $a \in A$; $T(s'|s,a): S \times A \times S \rightarrow [0,1]$ denotes the state transition function; $R(s,a): S \times A \rightarrow \mathbb{R}$ is the reward function; $\gamma \in [0,1]$ is the discount factor. We use $s^t$, $a^t$ and $r^t=R(s^t,a^t)$ to denote the state, action and reward at timestep $t$, respectively. Our goal is to learn a policy $\pi(a^t|s^t)$ that can maximize $\mathbb{E}[\hat{R}^t]$ where $\hat{R}^t=\Sigma^{H}_{k=t}\gamma^{k-t}r^k$ is the discounted return, and $H$ is the time horizon. Reinforcement learning \cite{sutton2018reinforcement} is a popular approach to solve the MDP problems. In practice, the environment can be noisy, so we can only get an observation $o^t$, which contains partial information of the state $s^t$. We have to learn the policy based on the observation history $[ o^{t-K}; ...; o^{t-1}; o^{t} ]$, which will be represented by $[o^k]^{t}_{k=t-K}$ for simplicity and $K$ is the history horizon. This setting is called partially-observable MDP (POMDP) \cite{spaan2012partially}. Given these settings and notations, we introduce our PDiT network as follows.

\subsection{Vanilla-PDiT}\label{sec:VanillaTIT}
\begin{figure}[t]
\centering
\includegraphics[width=1.0\columnwidth]{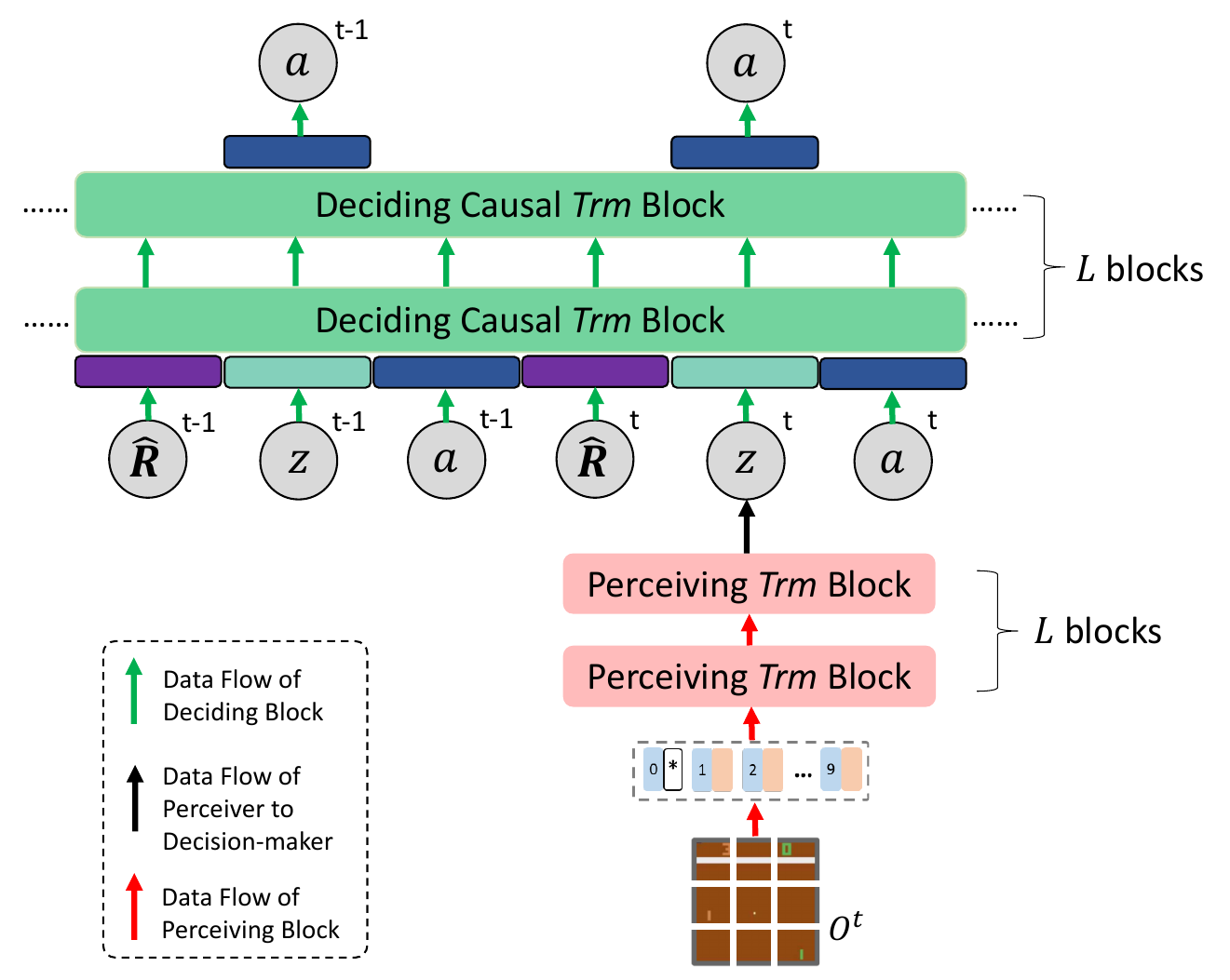}
\caption{The architecture of the proposed Vanilla-PDiT.}
\label{fig:Vanilla_TIT}
\end{figure}

Vanilla-PDiT is the minimal implementation of PDiT. As shown in Figure \ref{fig:Vanilla_TIT}, it stacks multiple perceiving Transformers to encode the multi-modal input and multiple deciding Transformers to learn the policy via the MDP decision sequences.

Specifically, the perceiving Transformer (which consists of $L$ perceiving blocks) focuses on \emph{the environmental perception} by processing the observation at the patch level, while the deciding Transformer (which is made up of $L$ deciding blocks) pays attention to \emph{the decision-making} (i.e., generating the action $a^t$) by conditioning on the history of the desired returns $[\hat{R}^k]^{t}_{k=t-K}$, the perceiver’s outputs $[z^k]^{t}_{k=t-K}$ and the actions $[a^k]^{t-1}_{k=t-K}$.

However, naively stacking the deciding Transformer on top of the perceiving one cannot always achieve the expected performance, as shown in our experiments. We suppose this is because there is no information interaction between the perceiving and deciding blocks, which may hinder the representation learning for reinforcement learning, as demonstrated by our feature visualization. So we propose the full PDiT to improve the performance. 

\subsection{PDiT}
\begin{figure}[t]
\centering
\includegraphics[width=1.0\columnwidth]{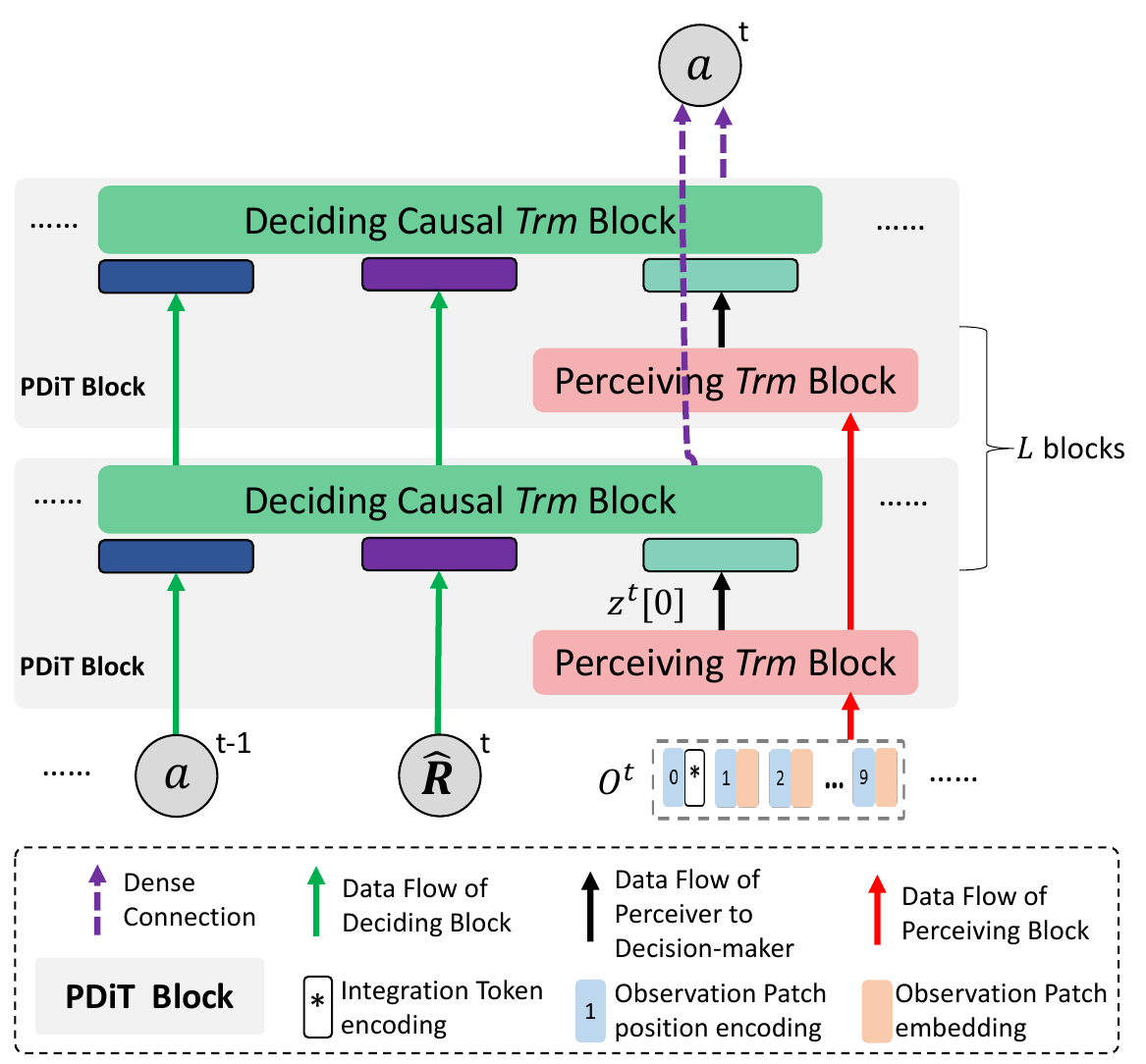}
\caption{The architecture of the proposed full PDiT, which stacks $L$ PDiT blocks (i.e., the grey rectangle). In each PDiT block, there is a perceiving block and a deciding block that are exactly the same as those of Vanilla-PDiT. Note that the perceiving blocks in the same layer share model parameters across different timesteps.}
\label{fig:Enhanced_TIT}
\end{figure}

As shown in Figure \ref{fig:Enhanced_TIT}, the full PDiT first builds a PDiT block by one perceiving block and one deciding block on the top, then stacks $L$ PDiT blocks to form the network. In this way, we \textbf{interleave} the perceiving and deciding Transformer blocks, such that the information can be fully fused in every PDiT block, while the perception and decision are still specialized by the perceiving and deciding blocks, respectively. In the following, we introduce the details. 

\begin{figure}[t]
\centering
\includegraphics[width=1.0\columnwidth]{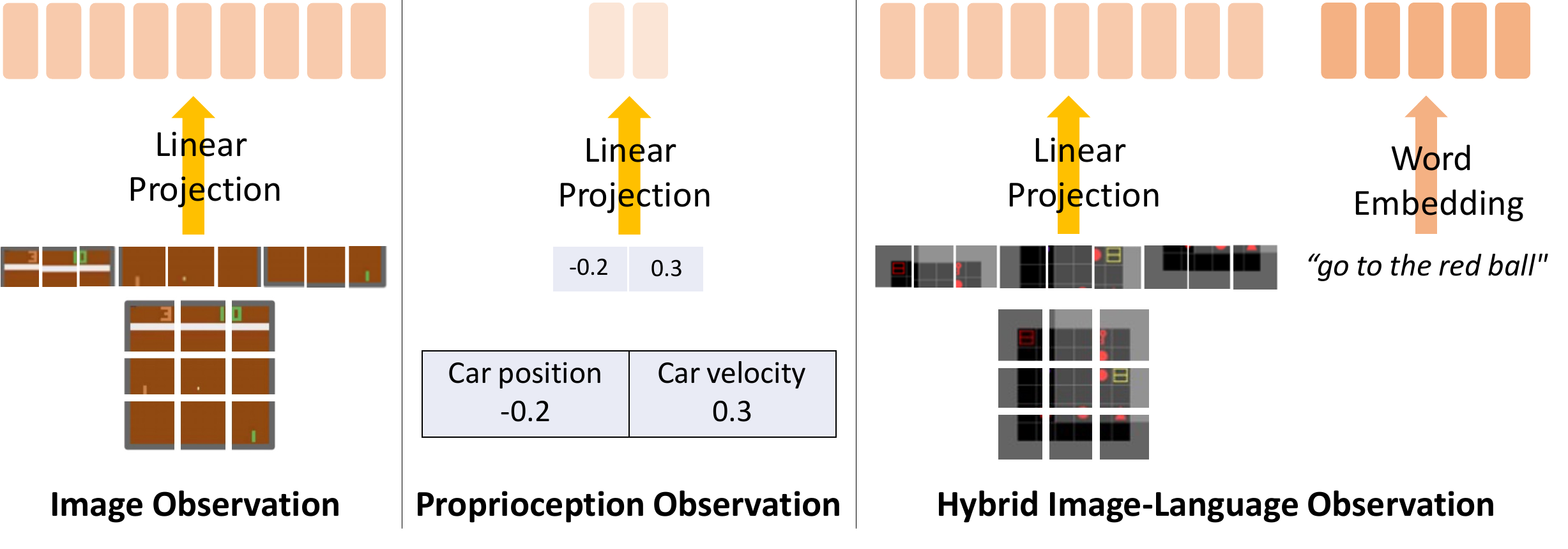}
\caption{The illustration of generating the observation patch embedding for different types of observations.}
\label{fig:ObservationPatchEmbedding}
\end{figure}

\subsubsection{\textbf{Dealing with the Multi-modal Observations}}
Different environments have different types of observations. As shown in Figure \ref{fig:ObservationPatchEmbedding}, we consider three common types: image observation in Atari, proprioception observation in MuJoCo, and hybrid image-language observation in BabyAI.

\textbf{(1) Transform into Sequential Input.} To feed multi-modal input into a Transformer, we will first ``sequentize'' the data. 

Given an \textbf{image observation} $o \in \mathbb{R}^{H \times W \times C}$, we split it into a sequence of \emph{observation patches} $o^p \in \mathbb{R}^{N \times (P^2 \times C)}$, where $(H, W)$ is the resolution of the original image observation; $C$ is the number of channels; $(P, P)$ is the resolution of each patch; and $N = HW/P^2$ is the resulting number of patches, which is known as the context length for the perceiving Transformer. After getting the observation patches, we map each patch $o^p_i$ into the \emph{observation patch embedding} with a trainable linear projection $E^p$:
\begin{eqnarray}
\hat{z}_0 = [o^p_1 E^p; o^p_2 E^p; ...; o^p_N E^p] \in \mathbb{R}^{N \times D^p}
\end{eqnarray}
where $E^p \in \mathbb{R}^{(P^2C) \times D^p}$, and $D^p$ is the dimension of the observation patch embedding.

Given a \textbf{proprioception observation} with $D$ entries, i.e., $o \in \mathbb{R}^{D}$, we take each entry as an observation patch independently: the sequence of observation patches will be $o^p \in \mathbb{R}^{N \times 1}$ where $N$ is equal to $D$. This is convenient and reasonable since each entry usually has atomic semantics, e.g., in the classic Mountain Car environment, the 2 entries of observation represent the atomic meaning of ``Car position'' and ``Car velocity'', respectively. If there is prior knowledge, we can generate the observation patches in other ways (e.g., several entries as a patch). Then, we again use a trainable linear projection $E^p \in \mathbb{R}^{1 \times D^p}$ to map the observation patches into the observation patch embeddings.

Given a \textbf{hybrid image-language observation}, the image will be processed as above, and the language will be processed using Word Embedding \cite{kusner2015word}. We concatenate them and get a total number of $N = HW/P^2 + N_w$ observation patch embeddings, where $N_w$ is the number of words in the language. Our method is motivated by Vision-Language Pre-training \cite{kim2021vilt}.

\textbf{(2) Integration Token Encoding.}
After getting the observation patch embeddings, a trainable \textit{integration token} $z_0^{integration} \in \mathbb{R}^{1 \times D^p}$ is added:
\begin{eqnarray}
\check{z}_0 = [z_0^{integration}; \hat{z}_0] \in \mathbb{R}^{(1+N) \times D^p}
\end{eqnarray}

\textbf{(3) Observation Patch Position Encoding.} 
We then add the \textit{observation patch position encoding}, which is a trainable parameter $E^p_{pos}$, to retain positional information:
\begin{eqnarray}
z_0 = \check{z}_0 + E^p_{pos}  \;  \;  \;  \;  \;  \;  E^p_{pos}  \in \mathbb{R}^{(1+N) \times D^p}
\end{eqnarray}
The resulting $z_0$ serves as the input of the perceiving Transformer.

\subsubsection{\textbf{Interleaving Perceiver and Decision Maker}} Now we introduce the PDiT block.

\textbf{(1) Perceiving Transformer Block.} 
The concrete operations in the $l$-th perceiving block $z_l = \textit{P-Trm}_{l}(z_{l-1})$ are as follows:
\begin{eqnarray}
\tilde{z}_l &=& z_{l-1} + \text{MSA}(\text{LN}(z_{l-1})) \;\;\;\;\;\; l = 1, ..., L \\
z_l &=& z_{l-1} + \text{FFN}(\text{LN}(\tilde{z}_{l-1})) \;\;\;\;\;\;\; l = 1, ..., L
\end{eqnarray}
where MSA($\cdot$), LN($\cdot$), and FFN($\cdot$) stand for the multi-headed self-attention, layer normalization, and feed-forward network used in the original Transformer \cite{vaswani2017attention}. There is no mask in the MSA so that all patches in an observation can be used for good environmental perception. Moreover, the perceiving block can build the \emph{spatial relationships} among observation patches within an observation by computing interactions between two observation patches.

For any layer $l$, there are 1+$N$ tokens in $z_l$ (i.e., $z_l \in \mathbb{R}^{(1+N) \times D^p}$), and the integration token $z_l[0] \in \mathbb{R}^{1 \times D^p}$ will serve as the integrated representation of all tokens. 

\textbf{(2) Deciding Transformer Block.} 
As shown in Figure \ref{fig:Enhanced_TIT}, the input of the deciding block is the trajectory across $K$ timesteps. We use the superscript $t$ and the subscript $l$ to represent the timestep and the layer, respectively. The input of the $l$-th deciding block is:
\begin{eqnarray}
y_{l-1} = [ \hat{R}^{t-K}_{l-1}; z^{t-K}_{l}[0]; a^{t-K}_{l-1}; ...; a^{t-1}_{l-1}; \hat{R}^t_{l-1}; z_l^t[0] ]
\end{eqnarray}
The detailed operations in the $l$-th deciding block $y_l = \textit{D-Trm}_l (y_{l-1})$ can be shown as follows:
\begin{eqnarray}
\tilde{y}_l &=& y_{l-1} + \text{MSA}(\text{LN}(y_{l-1})) \;\;\;\;\;\; l = 1, ..., L \\
y_l &=& y_{l-1} + \text{FFN}(\text{LN}(\tilde{y}_{l-1})) \;\;\;\;\;\;\; l = 1, ..., L
\end{eqnarray}
There are causal masks in the MSA so that the deciding block can build the \emph{temporal relationships} among different trajectory tokens across $K$ timesteps, which is helpful for decision-making in the POMDP setting.

\textbf{(3) PDiT Block.} 
As shown in Figure \ref{fig:Enhanced_TIT}, the PDiT block combines one perceiving block and one deciding block. Formally, the operations in the $l$-th PDiT block are:
\begin{eqnarray}
z^k_l &=& \textit{P-Trm}_{l} (z^k_{l-1}) \;\;\;\;\;\;\;\;\; k = t-K, ..., t \label{eq:innerT}\\
y_{l-1} &=& [ \hat{R}^{t-K}_{l-1}; z^{t-K}_{l}[0]; a^{t-K}_{l-1}; ...; a^{t-1}_{l-1}; \hat{R}^t_{l-1}; z^t_l[0] ] \\
y_l &=& \textit{D-Trm}_{l} (y_{l-1}) 
\end{eqnarray}
In Eq. (\ref{eq:innerT}), the perceiving blocks in the same layer share model parameters across different timesteps, so the number of model parameters in PDiT is comparable to existing methods. % (see Appendix 5.1 for details). 

Note that for any layer $l$, there are 2+$3K$ tokens in $y_l$. This is because there are 3 tokens (i.e., $\hat{R}, z[0], a$) for each timestep before $t$ and $\hat{R}^t, z^t[0]$ for timestep $t$. We use the last token $y_l[-1]$ as the representation of all tokens, which is a conventional operation for causal Transformers. 

\textbf{(4) PDiT Backbone Network.} 
We stack $L$ PDiT blocks to form the full PDiT network, ensuring that the spatial-temporal information is fully fused for good representation and decision. Then, we apply a dense connection design to concatenate the last token $y_l[-1]$ from all PDiT blocks and apply an FFN upon the concatenation to generate the final action:
\begin{eqnarray}
a^t &=& \text{FFN}( [ y_1[-1]; y_2[-1]; ...; y_L[-1] ] )
\end{eqnarray}
We found that the dense connection is empirically helpful for performance in the experiments.

\subsubsection{\textbf{Why interleaving the perceiving and deciding blocks is better (than stacking all deciding blocks to the end)?}}  Possible reasons are as follows. First, the outer is to make decisions. Once it receives the output from the perceiving, it should directly take actions instead of further abstractly embedding the representations by stacking multiple deciding blocks. Second,  this interleaving design encourages the information interaction between the perceiving and deciding blocks (but not mixed or fused up). Third, stacking too deep transformer layers may also cause losing focus.

\begin{table}
\begin{center}
\caption{PPO, CQL, and DT cover a lot of RL settings.}
\label{tab:Coverage_of_PPO_CQL_DT}
\begin{tabular}{lccc}
\toprule
Setting & PPO & CQL & DT \\
\midrule
Online / Offline & $\surd$ / $\times$ & $\times$ / $\surd$ & - \\
On-Policy / Off-Policy & $\surd$ / $\times$ & $\times$ / $\surd$  & - \\
Policy Gradient / Q-Learning & $\surd$ / $\times$ & $\times$ / $\surd$  & - \\
RvS \cite{emmonsrvs} & - & - & $\surd$ \\
\bottomrule
\end{tabular}
\end{center}
\end{table}

\subsection{Training Methods} 
This paper focuses on designing networks that are generally applicable to different deep RL settings, so we restrain ourselves from modifying the RL training algorithms and use the existing algorithms to train PDiT. Specifically, we train PDiT with PPO \cite{John2017PPO} based on Stable-baseline3 \footnote{https://stable-baselines3.readthedocs.io/}, CQL \cite{kumar2020conservative} based on D3rlpy \footnote{https://d3rlpy.readthedocs.io/}, and Reinforcement Learning via Supervised Learning (RvS) \cite{emmonsrvs} based on the official Decision Transformer (DT) \footnote{https://github.com/kzl/decision-transformer/}. As shown by Table \ref{tab:Coverage_of_PPO_CQL_DT}, these algorithms cover a lot of RL settings. We refer the readers to their original papers for more details.

\begin{table*}[!th]
\begin{center}
\caption{The episode returns for online Atari environments (trained by PPO). For NatureCNN, we report numbers directly from the official Stable-baseline3; other numbers are run by us. Compared to other CNN-based and Transformer-CNN-mixed networks, PDiT achieves the highest average performance with five independent runs (improving ResNet by 15.0\%). Best result is in boldface, and the second best is underlined.}
\label{tab:OnlineResults_PPO}
\begin{tabular}{l|cccc|c}
\toprule
 & Breakout & MsPacman & Pong & SpaceInvaders & Average \\
\midrule
NatureCNN & \underline{398$\pm$33} & 1754$\pm$172 & \underline{20.989$\pm$0.105} & 960$\pm$425 & 783.25 \\
ResNet & 397$\pm$57 & 1807$\pm$405 & \textbf{21.000$\pm$0.000} & \underline{1700$\pm$511} & \underline{981.25} \\
ResNet + Transformer (i.e., Catformer) & 242$\pm$41 & 1579$\pm$461 & 19.980$\pm$0.139 & 1427$\pm$597 & 817.00 \\
ResNet + Transformer + LSTM + Gating (i.e., CoBERL) & 358$\pm$34 & \underline{2190$\pm$327} & 19.460$\pm$1.557 & 821$\pm$314 & 847.12 \\
PDiT (ours) & \textbf{411$\pm$68} & \textbf{2246$\pm$326} & 20.750$\pm$1.577 & \textbf{1837$\pm$168} & \textbf{1128.69} \\
% (398+1754+20.989+960)/4 = 783.25
% (397+1807+21+1700)/4 = 981.25
% (242+1579+19.980+1427)/4 = 817.00
% (358+2190+19.460+821)/4 = 847.12
% (411+2246+20.750+1837)/4 = 1128.69
\bottomrule
\end{tabular}
\end{center}
\end{table*}

\begin{table*}[!th]
\begin{center}
\caption{The episode returns for offline MuJoCo (trained by RvS). We also include several concurrent methods (i.e., StARformer and GDT). DT and GDT results are reported from original papers; StARformer are taken from GDT; GATO is run by us. PDiT achieves the highest average performance over five runs (beat DT, StARformer, and GDT by 6.4\%, 40.7\%, and 2.8\%).}
\label{tab:OfflineResults_DT}
\begin{tabular}{ll|ccc||cc}
\toprule
Dataset & Environment Name & DT & GATO & PDiT (ours) & StARformer & GDT \\
\midrule
\multirow{3}{*}{Medium} & Halfcheetah & 42.6$\pm$0.1 & \underline{42.9$\pm$1.7} & 42.8$\pm$2.3 & \textbf{42.9$\pm$0.1} & \textbf{42.9$\pm$0.1} \\
& Hopper & \underline{67.6$\pm$1.0} & 58.7$\pm$2.5 & \textbf{68.2$\pm$2.4} & 59.5$\pm$4.2 & 65.8$\pm$5.8 \\
& Walker2d & 74.0$\pm$1.4 & \textbf{77.8$\pm$1.2} & \underline{77.6$\pm$0.6} & 73.8$\pm$3.5 & \textbf{77.8$\pm$0.4} \\
\midrule
\multirow{3}{*}{Medium-Replay} & Halfcheetah & 36.6$\pm$0.8 & 36.9$\pm$3.8 & \textbf{40.8$\pm$2.3} & 36.8$\pm$3.3 & \underline{39.9$\pm$0.1} \\
& Hopper & \underline{82.7$\pm$7.0} & 33.8$\pm$4.3 & \textbf{89.6$\pm$2.7} & 29.2$\pm$4.3 & 81.6$\pm$0.6 \\
& Walker2d & 66.6$\pm$3.0 & 64.7$\pm$0.8 & \underline{74.1$\pm$0.6} & 39.8$\pm$5.1 & \textbf{74.8$\pm$1.9} \\
\midrule
\multicolumn{2}{c|}{Average-normalized} & 61.68 & 52.47 & \textbf{65.52} & 47.00 & \underline{63.80} \\
% (42.6+67.6+74.0+36.6+82.7+66.6)/6 = 61.68
% (42.9+58.7+77.8+36.9+33.8+64.7)/6 = 52.47
% (42.8+68.2+77.6+40.8+89.6+74.1)/6 = 65.52
% (42.9+59.5+73.8+36.8+29.2+39.8)/6 = 47.00
% (42.9+65.8+77.8+39.9+81.6+74.8)/6 = 63.80
\multicolumn{2}{c|}{Average-original} & 4065.94 & 3443.94 & \textbf{4325.27} & 3074.53 & \underline{4209.11} \\
% mean-random: (-280.18-20.27+1.63)/3 = -99.61
% mean-expert: (12135.0+3234.3+4592.3)/3 = 6653.87
% 100*(original + 99.61)/(6653.87 + 99.61) = 61.68  ==> original = 4065.94
% 100*(original + 99.61)/(6653.87 + 99.61) = 52.47  ==> original = 3443.94
% 100*(original + 99.61)/(6653.87 + 99.61) = 65.52  ==> original = 4325.27
% 100*(original + 99.61)/(6653.87 + 99.61) = 47.00  ==> original = 3074.53
% 100*(original + 99.61)/(6653.87 + 99.61) = 63.80  ==> original = 4209.11
\bottomrule
\end{tabular}
\end{center}
\end{table*}

\begin{table*}[!th]
\begin{center}
\caption{The episode returns for offline MuJoCo (trained by CQL). We also compare other offline RL algorithms, i.e., BEAR \cite{kumar2019stabilizing}, BRAC-v \cite{wu2019behavior} and AWR \cite{peng2019advantage}. Here, CQL-MLP numbers are reported from the original paper; BEAR, BRAC-v and AWR are reported from the D4RL paper \cite{fu2020d4rl}. Compared to baselines, PDiT achieves the highest performance (improving MLP by 42.5\%).}
\label{tab:OfflineResults_CQL}
\begin{tabular}{ll|cc||cccc}
\toprule
Dataset & Task Name & CQL-MLP & PDiT (ours) & BEAR & BRAC-v & AWR \\
\midrule
\multirow{3}{*}{Medium} & Halfcheetah & \underline{44.4} & 42.6$\pm$0.1 & 41.7 & \textbf{46.3} & 37.4 \\
& Hopper & \underline{58.0} & \textbf{100.8$\pm$0.2} & 52.1 & 31.1 & 35.9 \\
& Walker2d & 79.2 & \textbf{84.1$\pm$0.9} & 59.1 & \underline{81.1} & 17.4 \\
\midrule
\multirow{3}{*}{Medium-Replay} & Halfcheetah & 46.2 & \textbf{47.8$\pm$0.1} & 38.6 & \underline{47.7} & 40.3 \\
& Hopper & \underline{48.6} & \textbf{99.2$\pm$1.7} & 33.7 & 0.6 & 28.4 \\
& Walker2d & \underline{26.7} & \textbf{53.6$\pm$3.7} & 19.2 & 0.9 & 15.5 \\
\midrule
\multicolumn{2}{c|}{Average-normalized} & \underline{50.52} & \textbf{71.35} & 40.73 & 34.62 & 29.15 \\
% (44.4+58.0+79.2+46.2+48.6+26.7)/6 = 50.52
% (42.6+100.8+84.1+47.8+99.2+53.6)/6= 71.35
% (41.7+52.1+59.1+38.6+33.7+19.2)/6= 40.73
% (46.3+31.1+81.1+47.7+0.6+0.9)/6= 34.62
% (37.4+35.9+17.4+40.3+28.4+15.5)/6= 29.15
\multicolumn{2}{c|}{Average-original} & \underline{3312.25} & \textbf{4719.00} & 2651.08 & 2238.44 & 1869.03 \\
% mean-random: (-280.18-20.27+1.63)/3 = -99.61
% mean-expert: (12135.0+3234.3+4592.3)/3 = 6653.87
% 100*(original + 99.61)/(6653.87 + 99.61) = 50.52  ==> original = 3312.25
% 100*(original + 99.61)/(6653.87 + 99.61) = 71.35  ==> original = 4719.00
% 100*(original + 99.61)/(6653.87 + 99.61) = 40.73  ==> original = 2651.08
% 100*(original + 99.61)/(6653.87 + 99.61) = 34.62  ==> original = 2238.44
% 100*(original + 99.61)/(6653.87 + 99.61) = 29.15  ==> original = 1869.03
\bottomrule
\end{tabular}
\end{center}
\end{table*}

\begin{table}[!th]
\begin{center}
\caption{The episode returns for offline BabyAI tasks.}
\label{fig:OfflineResults_BabyAI}
\begin{tabular}{l|ccc}
\toprule
 & DT & GATO & PDiT (ours) \\
\midrule
GoToRedBall & 0.969$\pm$0.01 & \underline{0.985$\pm$0.00} & \textbf{0.994$\pm$0.00} \\
GoToLocal & 0.884$\pm$0.02 & \underline{0.923$\pm$0.03} & \textbf{0.991$\pm$0.00} \\
PickupLoc & 0.719$\pm$0.02 & \underline{0.755$\pm$0.01} & \textbf{0.954$\pm$0.01} \\
PutNextLocal & 0.342$\pm$0.01 & \underline{0.435$\pm$0.02} & \textbf{0.881$\pm$0.01} \\
\midrule
Average & 0.729 & 0.775 & \textbf{0.955} \\
% (0.884+0.719+0.342)/3 = 0.648
% (0.923+0.755+0.435)/3 = 0.704
% (0.991+0.954+0.881)/3 = 0.942
% (0.969+0.884+0.719+0.342)/4 = 0.729
% (0.985+0.923+0.755+0.435)/4 = 0.775
% (0.994+0.991+0.954+0.881)/4 = 0.955
\bottomrule
\end{tabular}
\end{center}
\end{table}

\section{Experiment}
\subsection{Setting}\label{sec:Setting}
\subsubsection{Evaluation Environments.} We use Atari \footnote{https://www.gymlibrary.dev/environments/atari/}, MuJoCo\footnote{https://www.gymlibrary.dev/environments/mujoco/}, and BabyAI \footnote{https://minigrid.farama.org/environments/babyai/} as the evaluation environments. Specifically, Atari has image observations and discrete action spaces; MuJoCo has proprioception observations and continuous action spaces; while BabyAI has hybrid image-language observations and discrete action spaces. We consider these environments because they are credible benchmarks and their settings are diverse. % In the following, we only evaluate a subset scenarios of these environments due to limited computational resources and page constraints. We provide more results on other scenarios in the Appendix 4.

\subsubsection{Baseline Deep Networks.} Since PDiT is based on Transformers, we will mainly compare it with other Transformer-based networks. (1) Environment with image observations: we compare with NatureCNN \cite{mnih2015human}, ResNet \cite{RRL2021Shah}, ResNet + Transformer (i.e., Catformer \cite{davis2021catformer}), ResNet + Transformer + Gating + LSTM (i.e., CoBERL \cite{banino2021coberl}). (2) Environments with proprioception observations and hybrid observations: we compare with MLP, DT \cite{chen2021decision}, Transformer-based Behavior Cloning (i.e., GATO \cite{reed2022generalist}). 

\subsubsection{Hyperparameter Tuning.} To ensure fair comparison, we follow recent works \cite{chen2021decision,hu2023graph} and report results of most baselines from their original papers. We believe that these results are already tuned to the best by the original authors. For baselines run by us (only ResNet, Catformer, CoBERL and GATO), the common hyperparameters (e.g., training steps, learning rate, discount factor) are set as the open-source code repositories, while some specific hyperparameters (e.g., the number of network layers, embedding size) are tuned using the random grid-search as PDiT. % Specifically, ResNet, Catformer were tuned at most 10 times as PDiT, while CoBERL is twice of PDiT as its performance is unstable. % The details about training and evaluation are shown in the Appendix 2. 

\subsection{Result}\label{sec:Result}
\textbf{Training with PPO under Atari Environments.} The results are shown in Table \ref{tab:OnlineResults_PPO}. As can be seen, the conventional NatureCNN and ResNet can achieve satisfactory scores in most Atari environments. Compared to these conventional networks, ResNet + Transformer (i.e., Catformer) and ResNet + Transformer + LSTM + Gating (i.e., CoBERL) can achieve better average performance than NatureCNN but are worse than ResNet. It is because these mixed CNN-Transformer methods are unstable across different tasks, e.g., CoBERL gets a very high score in MsPacman, but the lowest score in SpaceInvaders. Compared to these baselines, PDiT achieves good performance in most testing cases.

\textbf{Training with RvS under MuJoCo Environments.} As observed in Table \ref{tab:OfflineResults_DT}, PDiT is better than DT and GATO, especially in practical datasets with complex distributions. Specifically, PDiT matches or exceeds DT and GATO by a small margin in ``Medium'' datasets generated from a single policy. In contrast, PDiT outperforms DT and GATO by a large margin in ``Medium-Replay'' datasets (also known as the ``Mixed'' datasets) that combine multiple policies. Since the mixed datasets with complex distributions are more likely to be common in practice, as mentioned in \cite{kumar2020conservative}, we expect that PDiT will work better than DT and GATO in practical applications.

\textbf{Training with CQL under MuJoCo Environments.} In the previous experiments, we find that training with RvS can sometimes result in small improvements. One possible reason may be that the ability of RvS to stitch offline data is not as strong as TD-learning. Thus, we test whether training with CQL, the state-of-the-art in offline TD-learning, can further improve PDiT. The results in Table \ref{tab:OfflineResults_CQL} show that PDiT can improve CQL-MLP by 42.5\% on average.

\textbf{Training with RvS under BabyAI Environments.} The results shown in Table \ref{fig:OfflineResults_BabyAI} demonstrate that PDiT is obviously better than DT and GATO under environments with hybrid image-language observations and sparse rewards.

\textbf{Summary:} All of the above results together prove that the proposed PDiT is generally applicable to a lot of RL settings, i.e., different RL training algorithms and evaluation environments with diverse characteristics. 

\textbf{More Results:} We provide more results on other classical scenarios and multi-task reinforcement learning in the Appendix 4. To highlight, we perform multi-task reinforcement learning with one compact network, our PDiT, without changing the network architecture. We use multiple tasks (e.g., GoToRedBall, GoToObj, and GoToSeq) to train PDiT, then evaluate PDiT on each task separately. PDiT with multi-task learning achieves a better average return than baselines with single-task learning.

\begin{figure}[th]
\centering
\includegraphics[width=1.0\columnwidth]{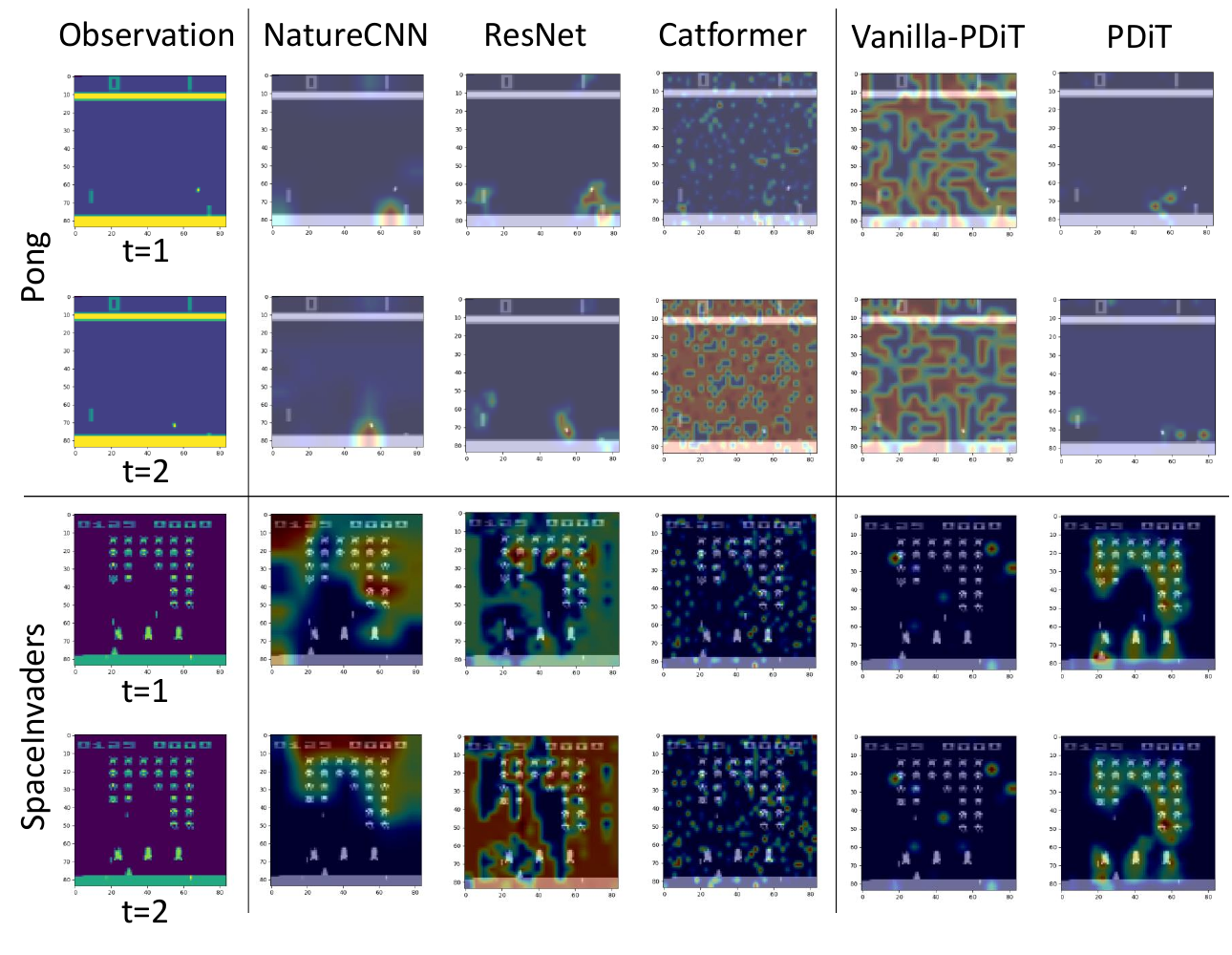}
\caption{The feature gradient visualization of 5 different methods. The first column shows the original observations randomly generated from Pong and SpaceInvaders: in the observations, objects need to be perceived.}
\label{fig:GradCAM}
\end{figure}

\subsection{Ablation Study}
We consider the following ablation networks:
\begin{enumerate}
    \item \textbf{Existing networks like DT}: There is one Transformer, which is worse than PDiT as shown by Table \ref{tab:OnlineResults_PPO} and \ref{tab:OfflineResults_DT}.
    \item \textbf{w/o Perceiver}: This removes the perceiving Transformer of PDiT, so only the deciding Transformer is used. In this network, the input of the deciding blocks is the \emph{observation embedding}, which is generated by linearly mapping the whole observation with a trainable parameter.
    \item \textbf{w/o Decision-maker}: This removes the deciding Transformer of PDiT, so only the perceiving Transformer is used. In this network, the $K$ observations are stacked in the same way as DQN and then processed by the perceiving blocks to generate the action.
    \item \textbf{Vanilla-PDiT}: There are two Transformers, but they are stacked naively as mentioned in Section \ref{sec:VanillaTIT}.
    \item \textbf{w/o Dense}: There are two Transformers, but the dense connection design is removed from the PDiT. In this network, only the output of the last PDiT block is used to generate the action (i.e., $a^t=\text{FFN}(y_L[-1] ]$).
\end{enumerate}

\begin{table*}[!th]
\begin{center}
\caption{The episode returns of ablation networks for online Atari environments (trained by PPO).}
\label{tab:AblationResults_PPO}
\begin{tabular}{ll|cccc|c|c}
\toprule
& & Breakout & MsPacman & Pong & SpaceInvaders & Average & Improve\\
\midrule
\multirow{2}{*}{One Transformer} & w/o Perceiver & 276$\pm$59 & 1588$\pm$516 & 19.350$\pm$2.441 & 1363$\pm$91 & 811.59 & -28.09\%\\
& w/o Decision-maker & 229$\pm$83 & 1591$\pm$396 & 20.180$\pm$2.034 & 1295$\pm$233 & 783.80 & -30.56\%\\
\midrule
\multirow{3}{*}{Two Transformers} & Vanilla-PDiT & 169$\pm$91 & 748$\pm$205 & 9.600$\pm$6.445 & 752$\pm$77 & 419.65 & \textbf{-62.82\%}\\
& w/o Dense & 121$\pm$34 & 1372$\pm$192 & 18.620$\pm$3.267 & 938$\pm$256 & 612.41 & \underline{-45.74\%}\\
& Full PDiT & 411$\pm$68 & 2246$\pm$326 & 20.750$\pm$1.577 & 1837$\pm$168 & 1128.69 & -\\
% (276+1588+19.35+1363)/4 = 811.59
% (229+1591+20.180+1295)/4 = 783.80
% (169+748+9.6+752)/4 = 419.65
% (121+1372+18.620+938)/4 = 612.41
\bottomrule
\end{tabular}
\end{center}
\end{table*}

\begin{figure*}[th]
\centering
\includegraphics[width=0.88\textwidth]{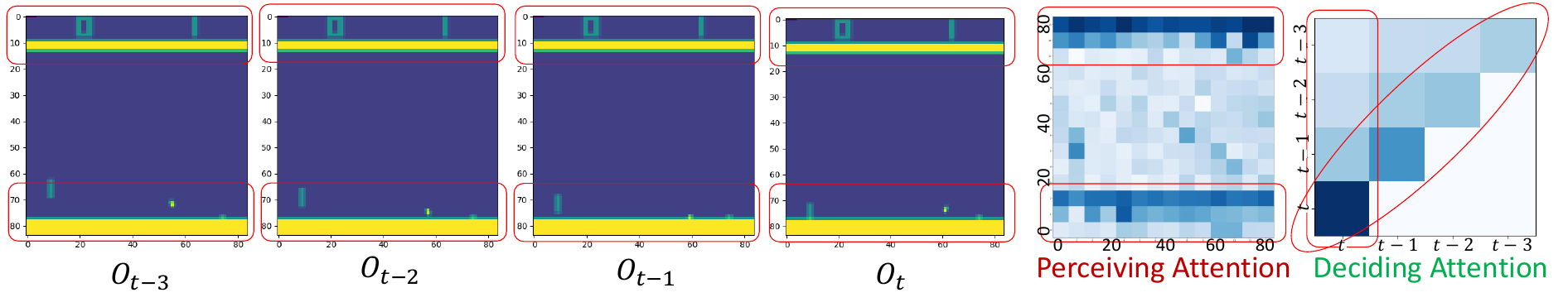}
\caption{The attention weights visualization of PDiT. Observations (i.e., the first four images) are randomly generated from Pong. The fifth and last images are the attention weights of the perceiving block and the deciding block, respectively.}
\label{fig:PongAttentionWeights}
\end{figure*}

The ablation results are shown in Table \ref{tab:AblationResults_PPO}. (1) Compared to PDiT, Vanilla-PDiT has the maximum performance degradation. It implies that although two Transformers are necessary for PDiT, naively combining them cannot get the expected performance, and the design of interleaving the perception and decision makes the most contribution. (2) Besides, ``w/o Dense'' has the second worst performance: this result verifies that the dense connection design is important. This is intuitive and reasonable since the dense network feeds all the outputs of each deciding Transformer to the final output, with more abundant information. As far as we know, previous studies have shown that the dense connection is also important for networks that are based on the CNN \cite{Huang2017Densely} and MLP \cite{Ota2020Can,sinha2020d2rl}, but PDiT is the first work to demonstrate this for purely Transformer-based networks. (3) Furthermore, both ``w/o Perceiver'' and ``w/o Decision-maker'' perform worse than PDiT, which indicates that both perceiving and deciding blocks are necessary for good performance.

\subsection{Visualization}
\subsubsection{\textbf{Feature Gradient Visualization}.}
We visualize the feature gradient of different methods by Grad-CAM \cite{GradCAM}, which is a typical method for visualization and explainability of deep networks. The results are shown in Figure \ref{fig:GradCAM}.

\emph{\textbf{From the spatial perspective, PDiT generates more explainable gradient maps than other methods.}} As can be seen in Figure \ref{fig:GradCAM}, the gradient of NatureCNN, ResNet, and PDiT is highly correlated with the objects in the observations in most cases. In contrast, Catformer and Vanilla-PDiT have disorganized gradient maps, and they sometimes generate unexplainable gradients, as shown by the Pong case. Furthermore, in the SpaceInvaders task, the gradient values are often positively-correlated with the density of the invaders (namely, the attention color is darker where there are many invaders that are close to the agent). 

\emph{\textbf{From the temporal perspective, PDiT generates more stable gradient maps than other methods.}} For example, when the observation changes slightly over a small time step, its gradient map does not change significantly. Thus, PDiT may learn consistent temporal representations for good decision-making, which may explain its stable performance across different tasks. In contrast, gradient maps of NatureCNN and Catformer have changed a lot.

\subsubsection{\textbf{Attention Weight Visualization}.}
In the above section, we have visualized the \emph{feature gradient} of different methods by \emph{Grad-CAM}. In this section, we visualize the \emph{attention weights} of the \emph{MSA operation} from the perceiving and deciding Transformers of our PDiT, as shown in Figure \ref{fig:PongAttentionWeights}. 

\emph{\textbf{The analyses from the spatial perspective.}} For the \emph{PDiT's perceiving block}: Here, we use the integration token as the key in the MSA operation because it serves as the integrated representation of all observation patches. As we can see, the observation image's top and bottom regions get higher attention weights since the top region is the game score, and the bottom region contains the paddles and balls, which are important for the game. In contrast, the central region of the observation image is mainly the background, which may have little influence on the game, so the attention weight values of the middle part are small. Based on this visualization, we conclude that the perceiving Transformer of PDiT can indeed capture important spatial information for good decision-making.

\emph{\textbf{The analyses from the temporal perspective.}} From \emph{the deciding block}'s perspective: Here, we use four observations to form the observation history to match the setting of the original DQN \cite{mnih2015human}. As we can see, the current observation $o_t$ usually gets the highest attention weights (i.e., the color on the diagonal is darker); moreover, the observation (e.g., $o_{t-1}$) close to the current timestep gets higher attention weight than the far away ones (e.g., $o_{t-3}$ and $o_{t-2}$); therefore the most updated information can be attended and used for decision-making. Based on this visualization, we conclude that PDiT (and the deciding Transformer) can indeed capture important temporal information for good decision-making.

% However, please note that the visualization is better used for explanation, but is not necessarily positively-correlated with the performance. For example, Catformer achieves good scores in Pong and SpaceInvaders, but its attention maps are disorganized, as shown in the main paper.

\section{Conclusion and Future Work}
This paper proposed the Perception and Decision-making Interleaving Transformer  (PDiT) network for deep RL. The key insight is cascading two specialized Transformers in a very natural way: the perceiving one processes the observations for good environmental perception, while the deciding one focuses on good decision-making. As far as we know, specializing perception and decision by a pure Transformer-based network is achieved for the first time. The experiments demonstrated that PDiT can achieve good results in many RL settings. Furthermore, PDiT can extract explainable representations from both spatial and temporal perspectives. 

With the goal to conceptually demonstrate the potential advantages of interleaving perception and decision-making for deep RL, we simply implement PDiT using the basic Transformer \cite{vaswani2017attention}. We expect that PDiT could further improve the performance with more advanced networks like Swin Transformer \cite{liu2021swin} and dedicated
optimization skills like contrastive learning \cite{eysenbach2022contrastive} (although this is not the focus of this paper). We give more discussions about the limitations and future directions in Appendix 5.

%%%%%%%%%%%%%%%%%%%%%%%%%%%%%%%%%%%%%%%%%%%%%%%%%%%%%%%%%%%%%%%%%%%%%%%%

%%% The acknowledgments section is defined using the "acks" environment
%%% (rather than an unnumbered section). The use of this environment 
%%% ensures the proper identification of the section in the article 
%%% metadata as well as the consistent spelling of the heading.

\begin{acks}
The authors would like to thank Dong Li, Jianye Hao and the anonymous
reviewers for their insightful comments. This work is supported by Basic Cultivation Fund project, CAS (JCPYJJ-22017), Strategic Priority Research Program of Chinese Academy of Sciences (XDA27010300), and the National Natural Science Foundation of China (Grants No. 62302189). % The contact authors are Hangyu Mao and Ziyue Li.
\end{acks}

%%%%%%%%%%%%%%%%%%%%%%%%%%%%%%%%%%%%%%%%%%%%%%%%%%%%%%%%%%%%%%%%%%%%%%%%

%%% The next two lines define, first, the bibliography style to be 
%%% applied, and, second, the bibliography file to be used.

\bibliographystyle{ACM-Reference-Format} 
\bibliography{sample}

%%%%%%%%%%%%%%%%%%%%%%%%%%%%%%%%%%%%%%%%%%%%%%%%%%%%%%%%%%%%%%%%%%%%%%%%

\end{document}